\documentclass[conference]{IEEEtran}
\IEEEoverridecommandlockouts
\usepackage{url}
\usepackage{cite}
\usepackage{amsmath,amssymb,amsfonts}
\usepackage{algorithmic}
\usepackage{graphicx}
\usepackage{textcomp}
\usepackage{xcolor}
\usepackage{graphicx}
\usepackage{caption}
\usepackage{lipsum}
\usepackage{hyperref}
\def\BibTeX{{\rm B\kern-.05em{\sc i\kern-.025em b}\kern-.08em
    T\kern-.1667em\lower.7ex\hbox{E}\kern-.125emX}}
\begin{document}

\title{Instant Photorealistic Neural Radiance Fields Stylization
}

\author{\IEEEauthorblockN{Shaoxu Li}
\IEEEauthorblockA{\textit{John Hopcroft Center for Computer Science} \\
\textit{Shanghai Jiao Tong University}\\
Shanghai, China \\
lishaoxu@sjtu.edu.cn}
\and
\IEEEauthorblockN{Ye Pan*}
\IEEEauthorblockA{\textit{John Hopcroft Center for Computer Science} \\
\textit{Shanghai Jiao Tong University}\\
Shanghai, China \\
whitneypanye@sjtu.edu.cn}
}

\twocolumn[{%
\renewcommand\twocolumn[1][]{#1}%
\maketitle
\begin{center}
    \centering
    \captionsetup{type=figure}
    \includegraphics[width=\textwidth]{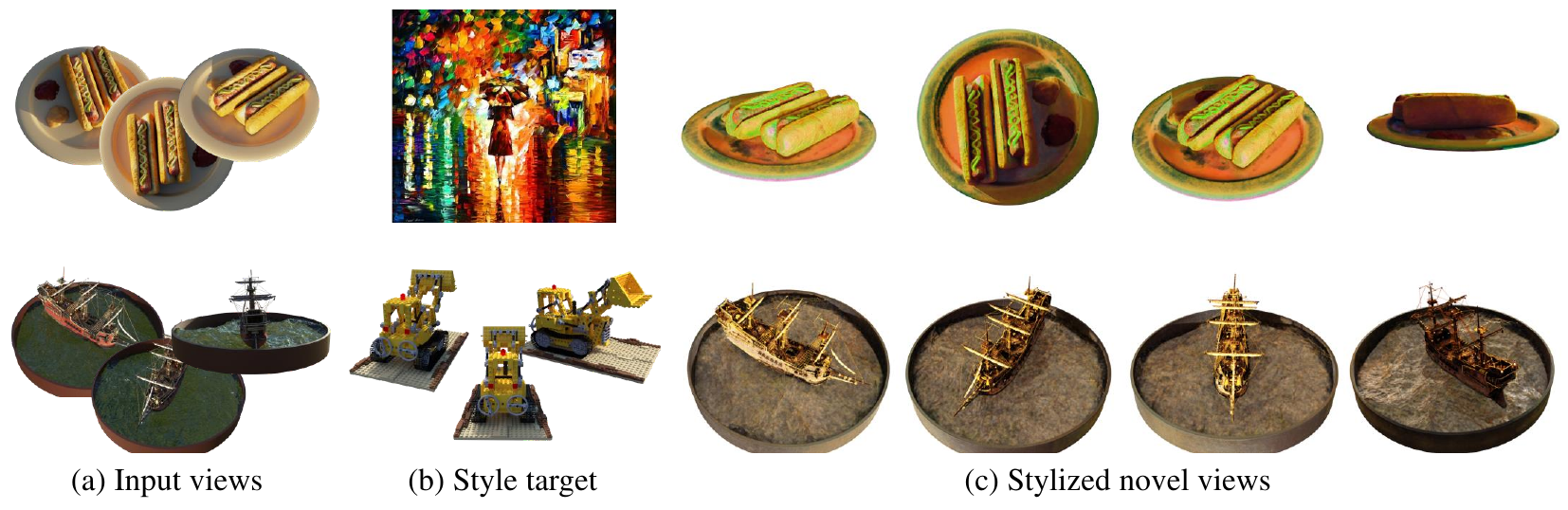}
    \captionof{figure}{Stylization results of our method. Given a set of images of 3D scenes (a) and a style target (b) (a style image or another set of images of 3D scenes),  our method is capable of generating stylized novel views (c) with a consistent appearance at various view angles.}
\end{center}%
}]


\begin{abstract}
We present Instant Neural Radiance Fields Stylization, a novel approach for multi-view image stylization for the 3D scene. Our approach models a neural radiance field based on neural graphics primitives, which use a hash table-based position encoder for position embedding. We split the position encoder into two parts, the content and style sub-branches, and train the network for normal novel view image synthesis with the content and style targets. In the inference stage, we execute AdaIN to the output features of the position encoder, with content and style voxel grid features as reference. With the adjusted features, the stylization of novel view images could be obtained. Our method extends the style target from style images to image sets of scenes and does not require additional network training for stylization. Given a set of images of 3D scenes and a style target(a style image or another set of 3D scenes), our method can generate stylized novel views with a consistent appearance at various view angles in less than 10 minutes on modern GPU hardware. Extensive experimental results demonstrate the validity and superiority of our method. Code: \href{https://github.com/lishaoxu1994/Instant-NeRF-Stylization} {\color{magenta}{https://github.com/lishaoxu1994/Instant-NeRF-Stylization}}
\end{abstract}

\section{Introduction}
Visual editing in 3D space has attracted increasing attention recently.  Given a set of images for a scene, novel view synthesis methods aim to generate high-quality images of the scene in arbitrary views.  In this paper, we focus on the stylization task of the scene with high quality. 
 
For the novel view synthesis task, Neural Radiance Fields (NeRF)\cite{mildenhall2020nerf} arouses great interest in researchers. NeRF has been a general 3D scene representation approach.  Many methods for kinds of tasks emerged with it, such as scene generation and editing\cite{Schwarz2020GRAF,Niemeyer2021GIRAFFE,Eric2021piGAN}, dynamic scene rendering\cite{park2021hypernerf,kania2022conerf}, large-scale scene rendering\cite{Tancik2022Block,li2022read}. Although high-quality results could be obtained with NeRF, many methods suffer from a long training time(more than one day). Some improvement methods tried to deal with this problem, such as Plenoxels\cite{yu2022Plenoxels}, TensorRF\cite{Chen2022TensoRF}. Instant-NGP\cite{mueller2022instant} is one of the state-of-the-art methods for acceleration. A typical NeRF embeds the position and direction inputs to do the rendering task. Instant-NGP uses multi-resolution hashing encoding for position embedding, accomplishing fast training with high quality. Trainable encoding parameters are arranged into $L$ levels, and conceptually stores feature vectors at the vertices of a grid. The position inputs are encoded through interpolation according to the relative position within its hypercube. The neural network uses encoded features and direction embedding results to synthesize the novel images. Our method uses this architecture to accomplish our stylization task.

Image stylizaton has been widely researched since \cite{Gatys_2015_Neural} presented a pioneering artistic style transfer algorithm. Fruitful follow-up works made it easier to adopt and higher quality for visualization\cite{Gu_2018_CVPR_Arbitrary,Kolkin_2019_CVPR_Style,Yao_2019_CVPR_Attention}. Video stylization is a similar task to 3D scene stylization. Video stylization aims to style images and keep consistency between different frames\cite{RuderDB_2016_Artistic,Chen_2017_Coherent,deng_2020_arbitrary}.

For the task of 3D scene stylization, the methods vary with the 3D scene representation approach, such as mesh\cite{Kato_2018_CVPR_Neural}, point cloud\cite{huang2021learning}, and NeRF\cite{Chiang2021Stylizing3S}. In this paper, we focus on the NeRF stylization, whose inputs are image sets of scenes and outputs are novel view stylzaiton images. Some recent works can transfer styles of NeRF scenes with consistency
\cite{Chiang2021Stylizing3S,Huang2022StylizedNeRF,chen2022upstnerf}. However, these methods made complex designs for composing 2D stylization methods, making them hard to extend. Besides, these methods suffer from a long training time. 

To this end, we present Instant Neural Radiance Fields Stylization, a novel approach for multi-view image stylization for the 3D scene. Our method leverages the multi-resolution hash table architecture for NeRF proposed by Instant-NGP\cite{mueller2022instant}. Our method uses the trainable feature vectors in the hash table to acquire the latent codes for stylization. First, we split the position encoder of instant neural graphics primitives into two parts: content and style. With the position encoder, our network can train for two scenes in less than 10 minutes. After training for normal scene synthesis, we execute AdaIN on position encoding features of novel view synthesis, with content and style voxel grid features as reference. Adjusted results are used for color predicting, and unadjusted results serve for density predicting. Unlike previous works whose style targets are only images, our method can accomplish stylization between two image sets for scenes. For stylization with images, we place the style images in the centre of the 3D space and treat them as if they were scenes.

Our main contributions are as follows:
\begin{itemize} 
\item We propose a novel instant neural radiance fields stylization method. The training for novel view stylization image synthesis only costs 10 minutes.

\item We propose, for the first time, style transfer from a 3D scene image set to a 3D scene image set. We extend the style target of NeRF stylization from images to 3D scene image sets.

\item We propose to manipulate the position embedding features(with position encoder) of NeRF for stylization. Our method can be extended to the migration of more image stylization methods.
\end{itemize}

\section{Related Work}
\noindent\textbf{Novel View Synthesis.} Novel view synthesis aims to generate high-quality novel view images for a scene, given a set of images. Recently, the neural radiance fields(NeRF)\cite{mildenhall2020nerf} aroused great interest in researchers. Follow-up works adjusted the details for more applications, such as scene generation and editing\cite{Schwarz2020GRAF,Niemeyer2021GIRAFFE,Eric2021piGAN}, dynamic scene rendering\cite{park2021hypernerf,kania2022conerf}, large-scale scene rendering\cite{Tancik2022Block,li2022read}.
Many methods are time-consuming(more than one day). Due to this, some methods have been proposed for acceleration. Plenoxels\cite{yu2022Plenoxels} used a sparse voxel grid with density and spherical harmonic coefficients at each voxel for scene synthesis, which achieves a speedup of two orders of magnitude compared to typical NeRF. TensorRF\cite{Chen2022TensoRF} considered the full volume field as a 4D tensor and proposed to factorize the tensor into multiple compact low-rank tensor components for efficient scene modelling. Instant-NGP\cite{mueller2022instant} presented a multi-resolution hash table of trainable feature vectors whose values are optimized through stochastic gradient descent. They implemented the whole system using fully-fused CUDA kernels, focusing on minimizing wasted bandwidth and compute operations. Instant-NGP achieved a speedup of several orders of magnitude and enabled high-quality rendering. Our method leverages this architecture for NeRF stylization.

\noindent\textbf{Image and Video Stylization.} Image stylization aims to synthesize an image with style characteristics extracted from style images. \cite{Gatys_2015_Neural} presented a pioneering artistic style transfer algorithm by optimizing the output image. Starting from \cite{Gatys_2015_Neural}, follow-up works adjusted the implement details for accessibility\cite{Liao_2017_Visual,Gu_2018_CVPR_Arbitrary}, quality\cite{Li_2016_CVPR_Combining,Kolkin_2019_CVPR_Style}, acceleration\cite{Johnson_2016_Perceptual,Yao_2019_CVPR_Attention},and application\cite{Yang_2017_CVPR_Awesome,Azadi_2018_CVPR_Multi}.

Video stylization aims to implement image style transfer with image frames from videos. Expect for the quality of style transfer, the consistency between different frames is another core target issue in video stylization. \cite{RuderDB_2016_Artistic} first presented an approach that transfers the style from one image to a whole video sequence with new initializations and loss functions. \cite{Chen_2017_Coherent} proposed the first end-to-end network for online video style transfer, which include an efficient network by incorporating short-term coherence and propagating short-term coherence to long-term. MCCNet\cite{deng_2020_arbitrary} fuses the exemplar style features and input content features for efficient style transfer while naturally maintaining the coherence of input videos.\cite{Gao_2020_WACV_Fast} designed a multi-instance normalization block (MIN-Block) to learn different style examples and a ConvLSTM module to encourage temporal consistency.

\begin{figure*}[ht]
\centering
\includegraphics[width=\linewidth]{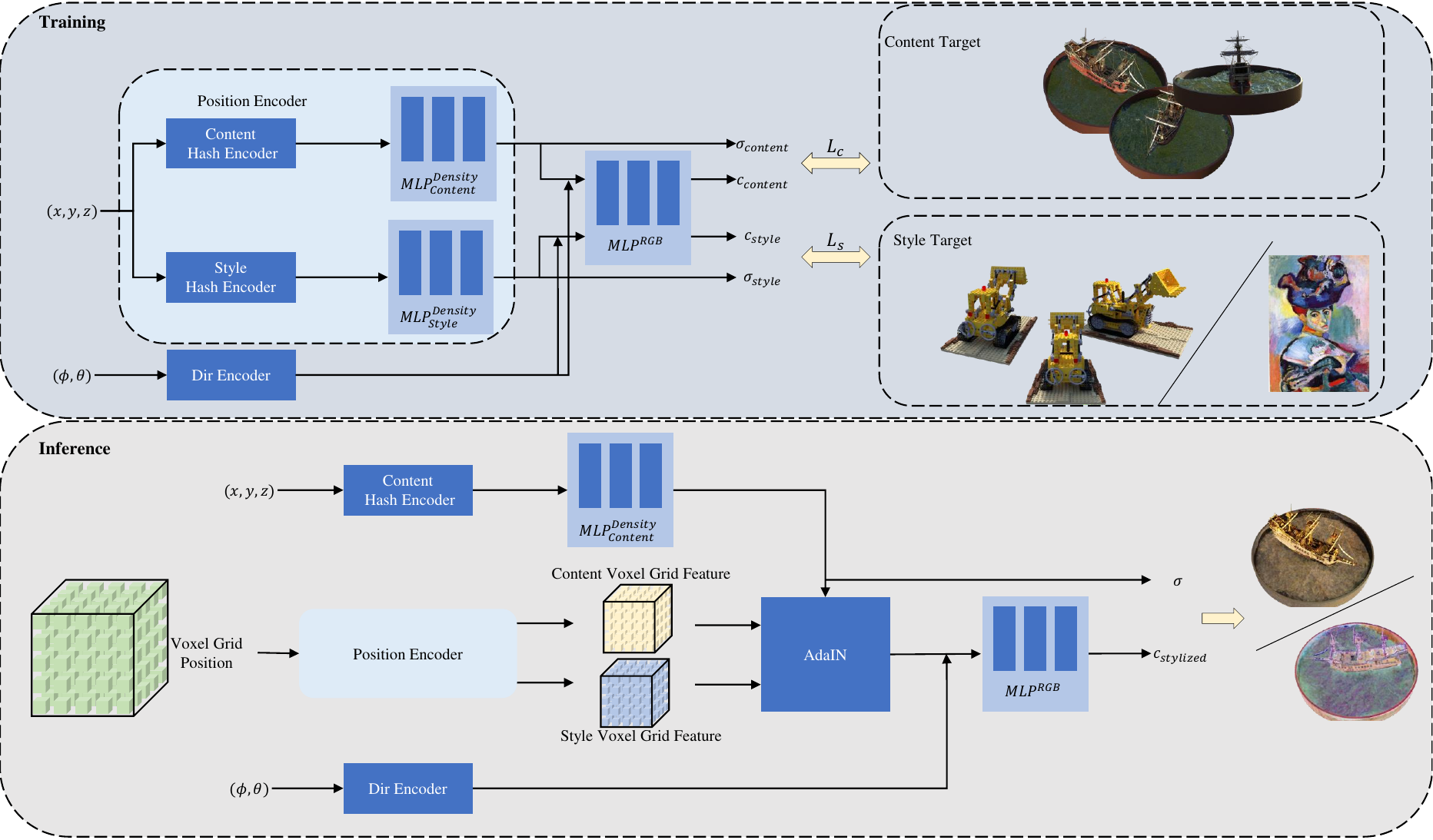}
\caption{The architecture of our method. Our approach uses a hash encoder for position embedding. We use two sub-position encoders for content and style images in the training process. Our style target could be a style image or another set of images of 3D scenes. For a style image, we treat it as if it is placed in the centre of 3D space. The two sub-position encoders share a $MLP^{RGB}$ for color calculation. The hash encoder promises the output of $MLP^{Density}$ as features, and shared $MLP^{RGB}$ promises the mixing of the features. We calculate the content and style feature for voxel grid positions in the inference stage. Then, AdaIN is executed for stylization, with these features for content and style mean and std parameters.}
\label{fig:model}
\end{figure*}

\noindent\textbf{3D scene style transfer.}3D scene style transfer aims to accomplish style transfer with 3D scene. These methods vary according to the 3D representation approach. 3D point cloud and mesh models are common. For example, NMR\cite{Kato_2018_CVPR_Neural} provided a Neural 3D Mesh Renderer (NMR), a renderer that includes 3D mesh editing operations such as 2D-to-3D style transfer.
\cite{Liu_2018_Paparazzi} constructed a differentiable triangle mesh renderer, which can backpropagate changes in the image domain to the 3D mesh vertex positions.
\cite{Mordvintsev_2018_Differentiable} focused on differentiable image parameterizations. Models with style textures could be obtained while the algorithm does not alter vertex positions.LSNV\cite{huang2021learning} constructed a single 3D representation, point cloud, for the holistic scene, and designed a point cloud transformation module to execute stylization.

NeRF is a popular method for 3D scene representation. Some methods emerge with NeRF 3D scene stylization. Style3D\cite{Chiang2021Stylizing3S} first attempted stylization for NeRF. Style3D\cite{Chiang2021Stylizing3S} used an implicit representation of the 3D scene with the neural radiance fields model and a hypernetwork to transfer the style information into the scene representation. StylizedNeRF\cite{Huang2022StylizedNeRF} proposed a mutual learning strategy for the stylized NeRF and 2D stylization method. StyleMesh\cite{Hollein_2022_StyleMesh} optimized an explicit texture for the reconstructed mesh of a scene and stylized it jointly from all available input images. ARF\cite{zhang2022arf} propose a novel deferred back-propagation method to enable optimization of memory-intensive radiance fields using style losses defined on full-resolution rendered images. SNeRF\cite{NguyenPhuoc2022SNeRFSN} alternates the NeRF and stylization optimization steps. INS\cite{fan2022unified} decouples the ordinary implicit function into a style implicit module and a content implicit module to encode the representations from the style image and input scenes separately. UPST\cite{chen2022upstnerf} proposed a hyper network to control the features of style images and use the 2D method to realize stylization.

\section{Method}
The overview of our method has been shown in Fig \ref{fig:model}. In the training process, we train a model with two branches for content and style encoding. Then, we calculate the content and style position encoding features for voxel grid positions in the rendering process. With these features, AdaIN\cite{huang2017adain} is executed for stylization. The style target can be an image set for a scene or one style image.

\noindent\textbf{Hash encoder with multilevel grid feature vectors.} For improving the training time of NeRF, many methods have been proposed. Instant-NGP\cite{mueller2022instant} is one of state-of-the-art methods.
Instant-NGP position encoder adopts a neural network with trainable weight parameters $\phi$ and trainable encoding parameters $\theta$. Encoding parameters are arranged into $L$ levels, and conceptually stores feature vectors at the vertices of a grid. Parameters in each level contain up to $T$ feature vectors with dimensionality. For a given input coordinate $(x,y,z)$, the encoder searches the surrounding voxels at $L$ resolution levels and the feature indices of these voxels by hashing their integer coordinates. For each level, the feature is the linear interpolation of the features according to the relative position of $(x,y,z)$ and surrounding voxels. The encoded features are a concatenation of the result of each level. Compared with typical position embedding, the hash encoder works faster, which could speed up from hours to minutes, ensuring high-quality rendering.

\subsection{Training with two parallel position encoders.}
As an implicit neural representation, NeRF takes 3D coordinates x=(x,y,z) and views direction d as inputs, colors c=(r,g,b), and volume density as outputs. Usually, the implicit function is multilayer perception(MLP). Our method uses the hash encoder with multilevel grid feature vectors, an efficient architecture proposed by Instant-NGP. Unlike NeRF for a single scene, our network serves for two scenes. Our network divides the position encoder(hash encoder and $MLP^{Density}$) for a single scene into two branches and keeps other parts unchanged. The two branches accomplish the position embedding for the content and style scenes in the training process. Moreover, the direction embedding is accomplished by a shared direction encoder(spherical encoder). We train the network for novel view synthesis with content and style target simultaneously, as shown in Fig\ref{fig:model}.

When fitting scenes, we use Huber loss for both content and style targets:
\begin{equation}
\label{equation:huber loss}
L=\begin{cases}
\frac{1}{2}rel^2, for\;rel<=\delta  \\
\delta(rel-\frac{1}{2}\delta),otherwise.
\end{cases}
rel=|C_{predict}-C_{target}|
\end{equation}
where $C_{predict}$, $C_{target}$ denotes rendering color value of images and the target value, $\delta$ is a hyperparameter. In the training process, The position encoder of content and style are optimized according to the content and style target respectively. The $MLP^{RGB}$ is optimized according to the content and style target together.

\begin{figure}[h]
\centering
\includegraphics[width=0.89\linewidth]{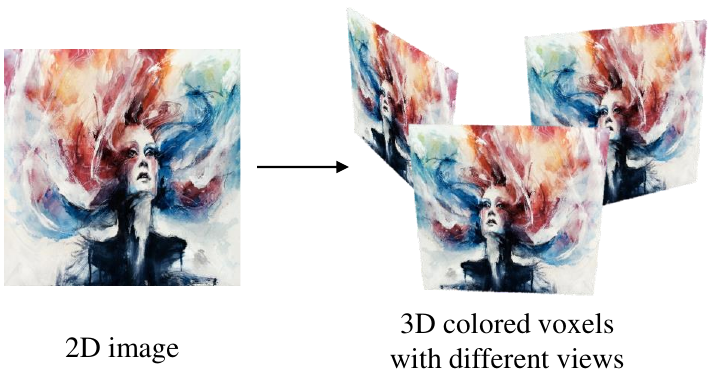}
\caption{2D style image and corresponding 3D colored voxels with different views.}
\label{fig:style in 3D}
\end{figure}

For image set styles, we treat it as a normal 3D scene. The content and style scene are trained jointly. For a style image, we treat it as if it is placed in the centre of the 3D space. We use 3D colored voxels to represent the 2D image pixels, as shown in Fig\ref{fig:style in 3D}.  Rendered 3D colored voxels images with different views are used for training like a typical 3D scene. In the training process, random views are selected for the optimization of the radiance field.

Compared with the network for one scene, our network has two position encoder sub-branches, which increases storage to a certain extent. The architecture of the hash encoder promises fast training for NeRF. Although we train two scenes meanwhile, our method's run time does not increase much. Our method could accomplish training in 10 minutes.

\begin{figure*}[t]
\centering
\includegraphics[width=\linewidth]{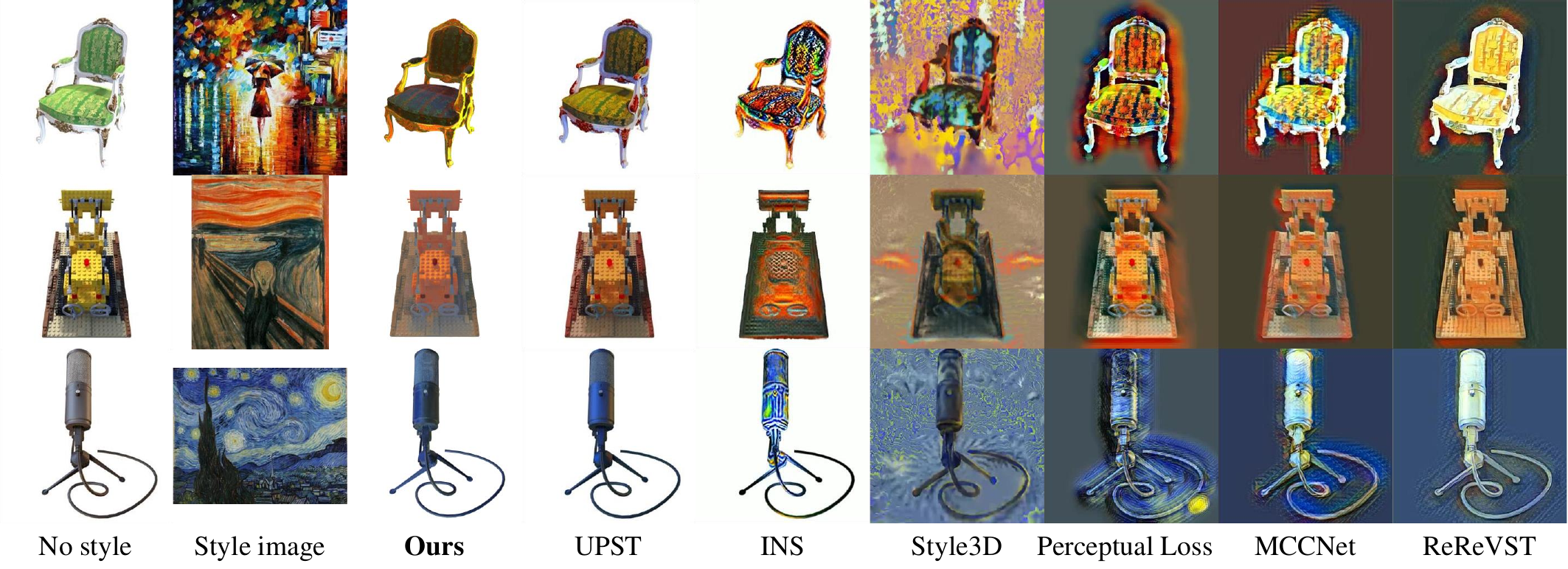}
\caption{Qualitative comparisons with artistic style images. We compare the stylization results of 3 scenes on the NeRF-Synthetic dataset.
Our method stylizes scenes with more precise geometry and competitive stylization quality.}
\label{fig:comp_1}
\end{figure*}

\begin{figure*}[t]
\centering
\includegraphics[width=\linewidth]{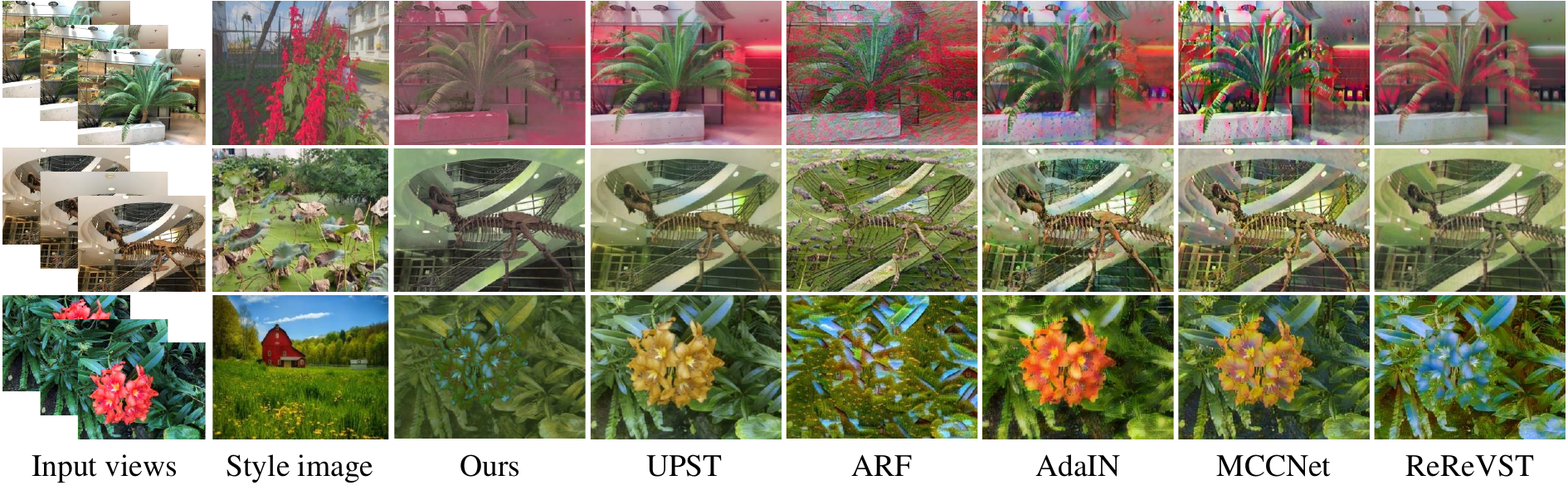}
\caption{Qualitative comparisons with photorealistic style images. We compare the stylization results of 3 scenes on the LLFF dataset. Our method stylizes scenes with more precise geometry and competitive stylization quality.}
\label{fig:comp_2}
\end{figure*}
\begin{figure*}[ht]
\centering
\includegraphics[width=\linewidth]{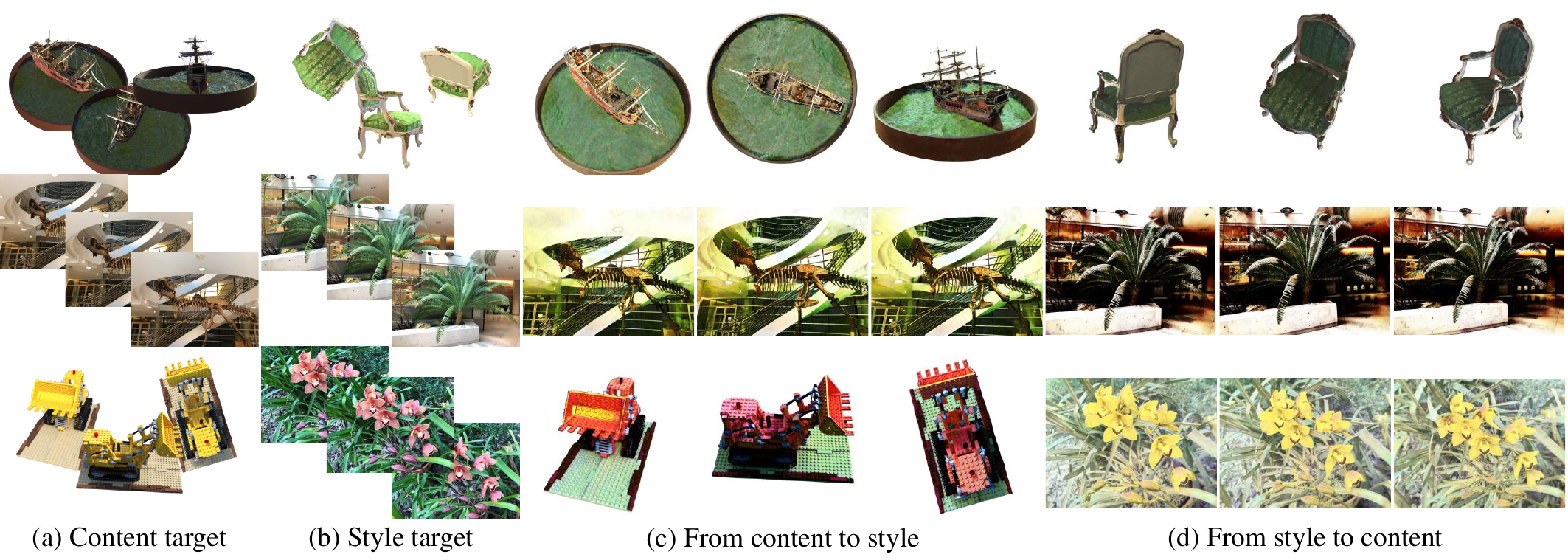}
\caption{Stylization results of our method on NeRF-Synthetic datasets and LLFF datasets. Given a set of images of 3D scenes (a) and a style target (b) (another set of images of 3D scenes),  our method is capable of generating stylized novel views (c)(from content to style) and (d)(from style to content) with a consistent appearance at various view angles.}
\label{fig:only_our}
\end{figure*}
\subsection{Stylization inference using AdaIN with voxel grid features.}
Once trained, our network can render high-quality images in novel views for content and style scenes. We use AdaIN\cite{huang2017adain} for stylization inference in the inference stage. We first calculate the results of the content and style position encoder, with voxel grid positions as input. We use voxel grid $V^{N_{V}\times N_{V}\times N_{V}}$ to represent the 3D space, where $N_{V}$ denotes voxel resolution. The results(Voxel Grid Feature) contain color and density features. 

To maintain the geometry consistency of the scene, the density is reserved for the final rendering. For color stylization, an AdaIN is executed. Unlike the original AdaIN\cite{huang2017adain}, our method executes AdaIN with reference parameters. The AdaIN module adjusts the feature $f(x,y,z)$ for stylization using Eq.\ref{equation:AdaIN}.
\begin{equation}
\label{equation:AdaIN}
AdaIN(f(x,y,z))= F_{\sigma}^{s}(\frac {f(x,y,z)-F_{\mu}^{c}}{F_{\sigma}^{c}})+F_{\mu}^{s}
\end{equation}
where $F_{\mu}^{c}$, $F_{\sigma}^{c}$, $F_{\mu}^{s}$ and $F_{\sigma}^{s}$ denote the mean and std of content and style voxel grid features. The AdaIN module adjusts the content position encoder features from the content distribution to the style target. According to the type of style target, our method can realize style transfer with artistic style images, photorealistic style images, and image set style images.

With AdaIN, the features are adjusted from content to style. In turn, the adjustment works from style to content features. Shared $MLP^{RGB}$ promises the transfer is reasonable. Our method uses the feature outputs of the position encoder as latent codes for 3D scene stylization. Our method can be extended to more 2D stylization methods. 

\begin{figure*}[ht]
\centering
\includegraphics[width=\linewidth]{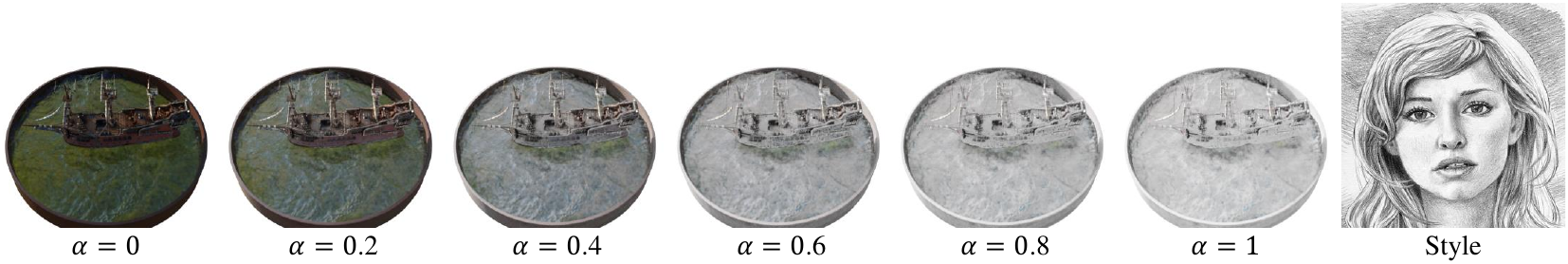}
\caption{Content-style trade-off. we can control the balance between content and style by changing the weight $\alpha$ in Eq.\ref{equation:AdaIN_control}}
\label{fig:control}
\end{figure*}
\section{Experiments Results and Analysis}
We conduct qualitative and quantitative experiments. Comparisons between our method and state-of-art methods illustrate the superior of our method. In qualitative evaluation, we execute comparison on NeRF-Synthetic datasets\cite{mildenhall2020nerf} with artistic style images and Local Light Field Fusion(LLFF) datasets\cite{mildenhall2019llff} with photorealistic style images. We demonstrate the results of style transfer between image set of scenes. And we shows the results of controlling the degree of stylization. In quantitative evaluation, we calculate the warped LPIPS metric\cite{zhang2018perceptual} for short and long-term consistency. Furthermore, we conduct a user study on the LLFF datasets for stylization and consistency comparisons. The code is accomplished using Jittor\cite{hu2020jittor} on a single Nvidia 3080 GPU.

\subsection{Qualitative Results}
\noindent\textbf{Style transfer with artistic style images.} In Fig.\ref{fig:comp_1}, we demonstrate style transfer results with artistic style images. We compare our method with state of art methods UPST\cite{chen2022upstnerf}, INS\cite{fan2022unified}, Style3D\cite{Chiang2021Stylizing3S}, Perceptual Loss\cite{Johnson_2016_Perceptual}, MCCNet\cite{deng_2020_arbitrary} and ReReVST\cite{Wang_2020_ReReVST}. These comparison results are cited from \cite{chen2022upstnerf}. Only INS, UPST and our method have good performances in geometry. Other results could not maintain high-quality results. For example, the whole space of the chair is cluttered in the result of Style3D. Artifacts exist with chair in the stylization results of Perceptual Loss ,MCCNet and ReReVST.

\noindent\textbf{Style transfer with photorealistic style images.} In Fig.\ref{fig:comp_2}, we demonstrate style transfer results with photorealistic style images. We compare our method with state of art methods UPST\cite{chen2022upstnerf}, ARF\cite{zhang2022arf}, AdaIN\cite{huang2017adain}, MCCNet\cite{deng_2020_arbitrary}, ReReVST\cite{Wang_2020_ReReVST}. These comparison results are cited from \cite{chen2022upstnerf}. The scene's geometry is preserved well with our method, UPST and ReReVST. For color stylization, the details vary with methods. 

\noindent\textbf{Style transfer between image set of scenes.} Besides style images, our method is applicable to work with style transfer between image sets of two scenes. To the best of our knowledge, we are the first to propose style transfer between image sets of scenes. So we only demonstrate some results without comparison in Fig. \ref{fig:only_our}. In the inference process, our method can accomplish stylization from content to style and style to content.

\noindent\textbf{Content-style trade-off.} The degree of stylization can be controlled in the inference by adjusting the style weight $\alpha$ in Eq.\ref{equation:AdaIN_control}. As shown in Fig.\ref{fig:control}, the style of images smoothly transfers between content and style targets by changing $\alpha$ from 0 to 1.
\begin{equation}
\label{equation:AdaIN_control}
AdaIN_{c}(f(x,y,z))= (1-\alpha)f(x,y,z) + \alpha AdaIN(f(x,y,z))
\end{equation}

\subsection{Quantitative Results}
\textbf{Consistency Measurement.}
We measure the short and long-term consistency using the warped LPIPS metric\cite{zhang2018perceptual}. A view v is warped with the depth expectation estimated by NeRF. The score is formulated as:
\begin{equation}
E(O_{i},O_{j})=LPIPS(O_{i},M_{i,j},W_{i,j}(O_{j}))
\end{equation}
where $W$ is the warping function and M is the warping mask. Only pixels within the mask $M_{i,j}$ are taken for the calculation. Five scenes in the LLFF dataset are taken for comparison. We use 20 pairs of views for each scene and gap 5$(O_{i},(O_{i+5})$ and 15$(O_{i},(O_{i+15})$ for short and long-range consistency calculation. The comparisons of short and long-range consistency are shown in Tab. \ref{tab:table_short} and Tab. \ref{tab:table_long}, respectively. Our method outperforms other methods by a significant margin.

\begin{table}[h]
    \centering
    \begin{tabular}{c|ccccc}
    \hline
         Method & Fern&Flower&Horns&Orchids&Trex \\
    \hline
         AdaIN &0.0091 &0.0077 &0.0098 &0.0099 &0.0085\\
         MCCNet &0.0075 &0.0055 &0.0088 &0.0083 &0.0070\\
         ReReVST &0.0045 &0.0030 &0.0041 &0.0060 &0.0027\\
         ARF &0.0055 &0.0033 &0.0061 &0.0088 &0.0064\\
         UPST &0.0030 &\textbf{0.0024} &0.0034 &0.0041 &0.0025\\
    \hline
         Ours &\textbf{0.0023} &0.0025 &\textbf{0.0025} &\textbf{0.0037} &\textbf{0.0018}\\
    \hline
    \end{tabular}
    \caption{Short-range consistency. We compare the short-range consistency using warping error. The best performance is in bold.}
    \label{tab:table_short}
\end{table}

\begin{figure}[h]
\centering
\includegraphics[width=0.85\linewidth]{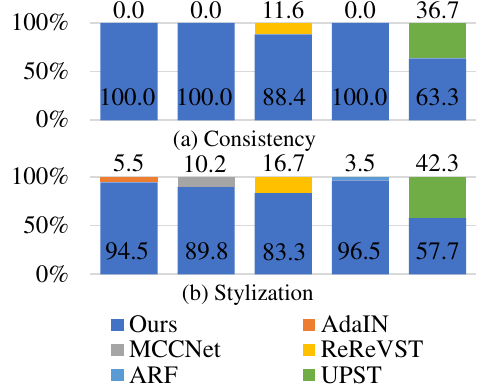}
\caption{User study. The number indicates the
percentage of preference.}
\label{fig:User study}
\end{figure}

\begin{table}[h]

    \centering
    \begin{tabular}{c|ccccc}
    \hline
         Method & Fern&Flower&Horns&Orchids&Trex \\
    \hline
         AdaIN &0.0497 &0.0458 &0.0493 &0.0353 &0.0212\\
         MCCNet &0.0407 &0.0217 &0.0270 &0.0471 &0.0168\\
         ReReVST &0.0195 &0.0110 &0.0220 &0.0394 &0.0285\\
         ARF &0.0355 &0.0185 &0.0247 &0.0397 &0.0218\\
         UPST &0.0320 &0.0100 &0.0157 &\textbf{0.0043} &0.0186\\
         \hline
         Ours &\textbf{0.0049} &\textbf{0.0099} &\textbf{0.0057} &0.0071 &\textbf{0.0140} \\
         \hline
    \end{tabular}
    \caption{Long-range consistency. We compare the long-range consistency using warping error. The best performance is in bold.}
    \label{tab:table_long}
\end{table}
\noindent\textbf{User study.}
A user study is conducted to compare the stylization and consistent quality of our method and others. We use the LLFF dataset for the study. Each scene in the LLFF dataset is stylized with our and other methods. We invited 50 participants and showed them the videos of stylization novel view synthesis results. For the same scene, we asked the participants to select the better results from our and another method, considering indicators stylization quality and geometry consistency. We collected 1000 votes for each indicator. Fig.\ref{fig:User study} show the result of the study. Our method outperforms other methods in both stylization quality and geometry consistency.

\subsection{Ablation Studies.} 
\begin{figure}[h]
\centering
\includegraphics[width=1.0\linewidth]{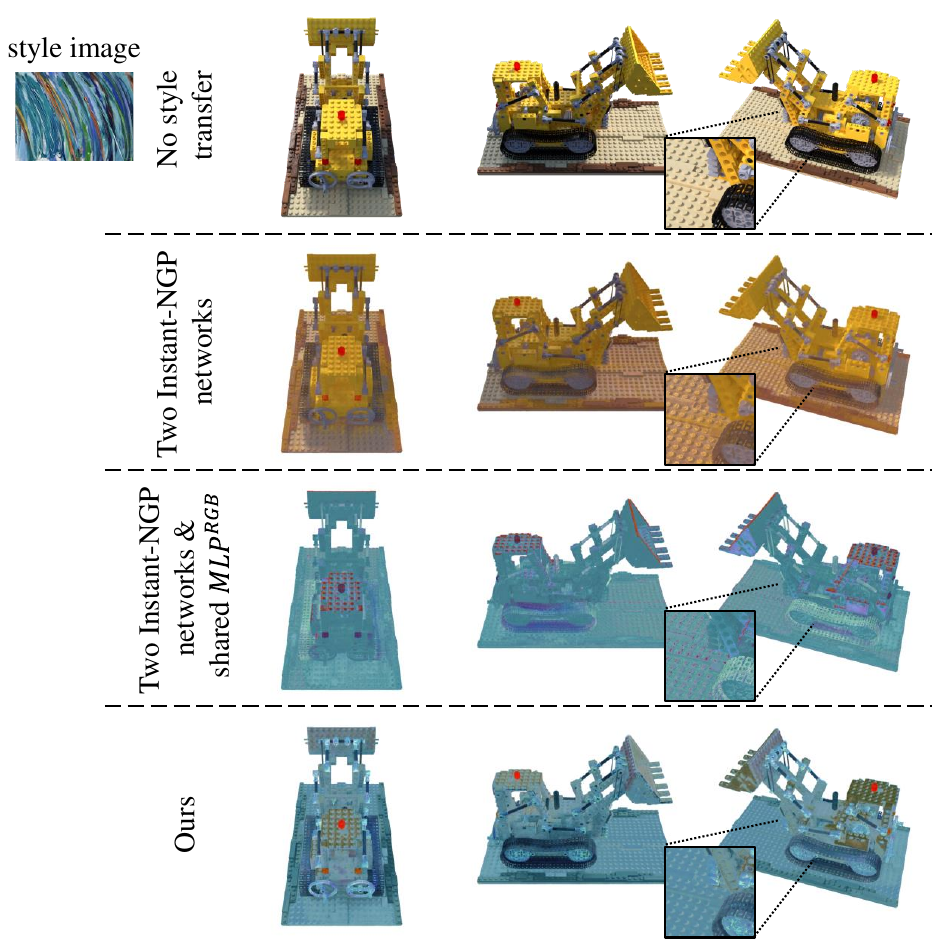}
\caption{Ablation study on two branches position encoder. We compare the results using two independent Instant-NGP networks, two independent Instant-NGP networks $\&$ shared $MLP^{RGB}$, and ours(two branches position encoder).}
\label{fig:ablation_two encoder}
\end{figure}

\noindent\textbf{Ablation study on two branches position encoder.}
We conduct an ablation study to demonstrate the necessity of using two branches networks. We compare the results of two individual
Instant NGP networks, two individual Instant NGP networks$\&$ shared $MLP^{RGB}$, and ours. For two individual Instant NGP networks, we train the two networks for content and style scenes, respectively. For two individual Instant NGP networks$\&$ shared $MLP^{RGB}$, we train the content scene first. Then we use fixed weights from $MLP^{RGB}$ of the content scene to train the style scene. The stylization inference procedure remains unchanged. Fig.\ref{fig:ablation_two encoder} shows the comparison results. With two individual Instant NGP networks, the color changes do not show characteristics from the style image. With two individual Instant NGP networks$\&$ shared $MLP^{RGB}$, the details of stylization results are unideal and contain much noise.

\begin{figure}[h]
\centering
\includegraphics[width=\linewidth]{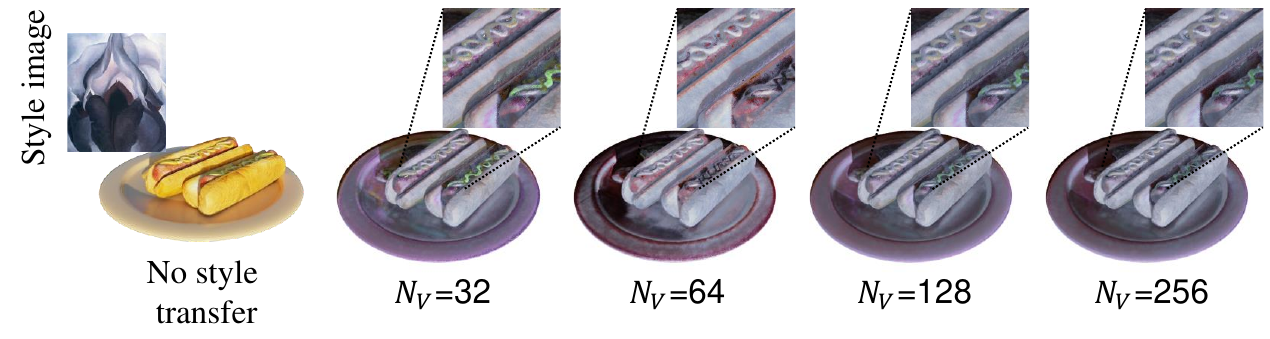}
\caption{Ablation study on two branches position encoder. We compare the results using two independent Instant-NGP networks, two independent Instant-NGP networks $\&$ shared $MLP^{RGB}$, and ours(two branches position encoder).}
\label{fig:ablation_voxel size}
\end{figure}

\noindent\textbf{Ablation study on voxel grid size.} We conduct an ablation study to decide the number of voxels. The results are presented in Fig.\ref{fig:ablation_voxel size}. The precision increases with the resolution of voxels $N_{V}$. However, higher-resolution voxels need more space, and the inference time increases. We choose $N_{V}=32,64,128,256$ to find the balance between inference quality and cost. More good results appear with the increase of $N_{V}$. From $N_{V}=128$ to $N_{V}=256$, the results change slightly. So we choose $N_{V}=128$ as our voxels resolution.

\begin{figure}[h]
\centering
\includegraphics[width=0.94\linewidth]{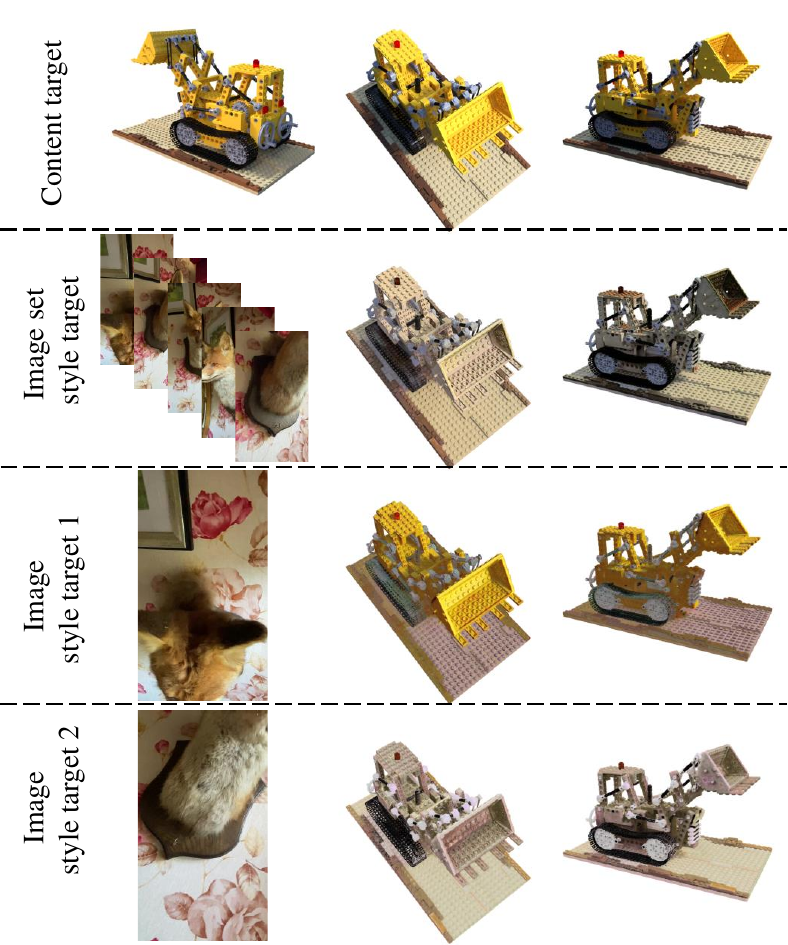}
\caption{Ablation study on image set style images. We compare the results using image set and single image as style target. The first line shows the content image set. The second line shows the results of using image set for style transfer. The last two lines show the results of using single image as the style target. The last two style target are randomly selected from the image set of the second line.}
\label{fig:ablation_image set}
\end{figure}
\noindent\textbf{Ablation study on image set style images.} We propose the first style transfer method from a 3D scene image set to a 3D scene image set. The style transfer between two scene could be accomplished with the image set style target or single image from the image set. We conduct an ablation
study to demonstrate the superiority of using image set style target. The results are presented in Fig.\ref{fig:ablation_image set}. The last two results only learns the characteristic from single image, which is far from the characteristic of the whole scene. Compared with using single image as style target, using image set is better for representing the characteristic of the 3D scene.

\section{Discussion}
While our method shows competitive quality and better speed compared to state-of-the-art NeRF stylization methods, there are still several challenges that need to be addressed in future work. The position and direction encoder output features are combined for color calculation. We only execute AdaIN on the position encoder features, leaving direction encoder features unchanged. So our method does not handle stylization on the illumination and reflection. 

Trainable encoding parameters are the core for fast scene training. For one scene, these parameters are random in twice training. Our method only promises the characteristics of stylizaiton results from content to style. Nevertheless, the details of the results vary with the random encoding parameters. For stable results, finding a solution for fixed $MLP^{RGB}$ for multi-scenes is necessary. If solved, the stylization could be accomplished with any two scenes using the fixed $MLP^{RGB}$.

\section{Conclusion}
In this paper, we present Instant Neural Radiance Fields Stylization, a novel approach that instantaneously styles 3D scenes. We split the position encoder of instant neural graphics primitives into two parts. With this architecture, our network can train neural radiance fields for two scenes in less than 10 minutes. Due to the output features of the position encoder accounting for scene synthesis, it could be used for scene edit. We use AdaIN for scene stylization with voxel grid features, which could be extended to more image stylization methods. Our method extends the style target from style images to image sets of scenes. Our method can generate stylized novel views with a consistent appearance at various view angles, given a set of images of 3D scenes and a style target(a style image or another set of images of 3D scenes). Extensive experimental results demonstrate the validity and superiority of our method.

\bibliographystyle{IEEEbib}
\bibliography{icme2023template}

\end{document}